%% file: main.tex
\documentclass[conference]{IEEEtran}
\IEEEoverridecommandlockouts

\usepackage{amsmath,amssymb,amsfonts,booktabs}
\usepackage{algorithm, algorithmic}
\usepackage{graphicx}
\usepackage{bm}
\usepackage{textcomp}
\usepackage{xcolor}
\usepackage{soul}
\usepackage{subcaption}
\usepackage{multirow}
\usepackage{siunitx}
\usepackage[stretch=40,shrink=40]{microtype}

\usepackage[normalem]{ulem}

\DeclareMathOperator*{\argmin}{argmin}

\DeclareMathOperator*{\topK}{topK}
\DeclareMathOperator*{\softmax}{softmax}

\newcommand{\compl}{\mathbb{C}}         
\newcommand{\real}{\mathbb{R}}          
\newcommand{\trans}{^{\text{T}}}		
\newcommand{\herm}{^{\text{H}}}			

\def\BibTeX{{\rm B\kern-.05em{\sc i\kern-.025em b}\kern-.08em
    T\kern-.1667em\lower.7ex\hbox{E}\kern-.125emX}}
\begin{document}

\title{Learning Structured Compressed Sensing with Automatic Resource Allocation\\
\thanks{This work was supported by the Thuringian Ministry of Economic Affairs, Science and Digital Society (TMWWDG).}
}

\author{\IEEEauthorblockN{Han Wang\textsuperscript{*,\ddag}, Eduardo P\'{e}rez\textsuperscript{*,\ddag}, Iris A. M. Huijben\textsuperscript{\dag}, Hans van Gorp\textsuperscript{\dag}, Ruud van Sloun\textsuperscript{\dag}, Florian R\"{o}mer\textsuperscript{*,\ddag}}
\IEEEauthorblockA{\textsuperscript{*}\textit{Applied AI Signal Processing and Data Analysis, Fraunhofer Institute for Nondestructive Testing IZFP}, Germany \\
\textsuperscript{\dag}\textit{Department of Electrical Engineering, Eindhoven University of Technology}, The Netherlands \\
\textsuperscript{\ddag}\textit{Electronic Measurements and Signal Processing, Ilmenau University of Technology}, Germany\\
Email: han.wang@izfp.fraunhofer.de}}

\maketitle

\begin{abstract}
Multidimensional data acquisition often requires extensive time and poses significant challenges for hardware and software regarding data storage and processing. Rather than designing a single compression matrix as in conventional compressed sensing, structured compressed sensing yields dimension-specific compression matrices, reducing the number of optimizable parameters. Recent advances in machine learning (ML) have enabled task-based supervised learning of subsampling matrices, albeit at the expense of complex downstream models. Additionally, the sampling resource allocation across dimensions is often determined in advance through heuristics. To address these challenges, we introduce Structured COmpressed Sensing with Automatic Resource Allocation (SCOSARA) with an information theory-based unsupervised learning strategy. SCOSARA adaptively distributes samples across sampling dimensions while maximizing Fisher information content. Using ultrasound localization as a case study, we compare SCOSARA to state-of-the-art ML-based and greedy search algorithms. Simulation results demonstrate that SCOSARA can produce high-quality subsampling matrices that achieve lower Cram\'er Rao Bound values than the baselines. In addition, SCOSARA outperforms other ML-based algorithms in terms of the number of trainable parameters, computational complexity, and memory requirements while automatically choosing the number of samples per axis.
\end{abstract}



\input{rough_intro}
\input{sec_scos}
\input{rough_ara}
\input{sec_optimization_target}

\input{sec_evaluation}
\input{sec_conclusion}

\clearpage
\bibliographystyle{unsrt}
\bibliography{reference}

\end{document}

%% file: rough_intro.tex
\section{Introduction}

\noindent Many practical problems involve multidimensional data with several axes that represent different domains \cite{dzemyda2013multidimensional}, e.g. space, time, Doppler shift, color channels, azimuth and elevation angles, etc. \cite{li2008mimo, vlaardingerbroek2013magnetic, holmes2005post}. 
Such data is information-rich, but quickly yields large data volumes and may involve time-consuming measurement procedures.

Compressed sensing (CS) \cite{eldar2012compressed} addresses the data volume and, if done judiciously, reduces the measurement time by measuring only a small number of linear projections of the data. The process of collecting these projections is represented through a rectangular matrix referred to as a compression matrix. Such compression matrices are classically designed as random matrices \cite{donoho2006compressed}. Such designs are optimal, but only asymptotically for large matrices or in expectation, and are also difficult to implement in hardware. This motivates the systematic design of optimal subsampling matrices, which is the goal of this paper. 

Subsampling matrices are a particular case of compression matrices with \emph{one-hot} rows, whereby only a subset of the original samples are measured \cite{lustig2007sparse, huijben2020learning, kirchhof2021frequency, wangius2023}. Although this results in loss of information when compared to random linear projections, the result is a CS scheme that is easily implemented through omission, planning, or reprogramming of the original measurement procedure without the need for additional hardware \cite{perez2020subsampling}. However, the design of subsampling matrices involves two combinatorial optimization subproblems: (I) ``choosing $k$ out of $n$ items'', and (II) ``distributing $k$ items into $q$ bins''. The present work addresses these problems on several fronts.


Problem (I) naturally arises when subsampling data. Advances in ML architectures and task-based learning have made it possible to use ML to learn a subsampling matrix\cite{huijben2020learning, wangicassp2024, mulleti2023learning, wang2023deep}. Due to the typically large volume of multidimensional data, task-based ML can be time-intensive or prohibitively resource-intensive. For example, a commonly-used deep unrolled neural network LISTA \cite{monga2021algorithm,gregor2010learning} grows linearly with the number of unrolled iterations and also scales with the product of the dimension sizes of the input data (e.g. transmitters $\times$ receivers $\times$ time domain samples $\times$ size of the region of interest). 

Problem (II) appears when structured CS of multidimensional data is applied \cite{wangicassp2024}, where each axis is compressed separately, turning a large subsampling problem into several smaller ones. More explicitly, the number of design parameters is reduced from the product of data dimension lengths to the sum of the data dimension lengths. However, the number of samples must now be distributed among the dimensions of the data. This resource allocation problem is often addressed through heuristics \cite{huijben2020learning, wang2023deep, wangicassp2024}. 

We alleviate the large-scale nature of problem (I) by replacing task-based learning with Fisher information maximization and exploiting the structure in the data. The Fisher Information Matrix (FIM) provides an alternative to task-based optimization \cite{kay1993fundamentals}, as commonly used in multistatic localization \cite{fatima2024optimal}. The FIM and its inverse, the Cram\'{e}r-Rao Bound (CRB), are known to accurately describe the performance of CS methods when the signal-to-noise ratio is high and quantization error is low \cite{blanchard2011compressed, austin2009relation, austin2013dynamic, chi2011sensitivity, ben2010cramer}. This makes the unsupervised maximization of the trace of the FIM an attractive alternative to task-based approaches in optimizing subsampling matrices. 
We address problem (II) by formulating the structured subsampling problem so that the number of samples per axis is learned automatically during training.

In this work, we propose a framework for the design of structured subsampling matrices by means of ML-based unsupervised maximization of the trace of the FIM. The framework, dubbed Structured COmpressed Sensing with Automatic Resource Allocation (SCOSARA), automatically distributes the samples among the axes of multidimensional data.

%% file: sec_scos.tex
\section{Structured subsampling}\label{sec_scs}

\noindent Data acquisition is performed across $q$ different dimensions, and the number of samples in the $i$th dimension is $N_i$ (with $i=1, 2, \dots, q$). The $q$-dimensional data array can be expressed as $\mathbf{Y} \in \mathbb{R}^{N_1 \times N_2 \times \dots \times N_q}$, which can be vectorized as $\mathbf{y}\in \mathbb{C}^{N_\Pi}$, with $N_{\Pi} = \Pi_{i=1}^qN_i$ the total number of samples. The sampling process is commonly represented as a linear transformation from the to-be-estimated signal $\mathbf{x}\in \compl^{N_{\rm{s}}}$ to this measurement:
\vspace{-0.1cm}
\begin{equation}\label{eq_forward}
    \mathbf{y=Ax+n},
\vspace{-0.1cm}
\end{equation}
where $\mathbf{n}\in \compl^{N_\Pi}$ represents circularly-symmetric Additive White Gaussian Noise (AWGN) following $\mathcal{CN}(\boldsymbol{0}, \boldsymbol{\Sigma})$ and $\mathbf{A}\in \compl^{N_\Pi\times N_{\rm{s}}}$ is the signal model. 


We introduce a
subsampling matrix $\mathbf{S} \in \{0,1\}^{M_\Pi \times N_\Pi}$ that selects $M_\Pi$ out of $N_\Pi$ data samples thanks to its row-wise one-hot structure, resulting in subsampled measurement $\mathbf{y}_{\rm{s}}\in \compl^{M_\Pi}$ through
\vspace{-0.1cm}
\begin{equation}\label{eq_cs}
    \mathbf{y}_{\rm{s}}=\mathbf{S}\mathbf{y}.
\vspace{-0.1cm}
\end{equation}

%

Structured COmpressed Sensing (SCOS) compresses data along each $i$th dimension separately. Structured subsampling -- the focus of this work -- is an instance of SCOS, where each $i$th axis is \textit{subsampled} separately. To this end, we introduce subsampling matrices $\mathbf{S}_i \in \{0,1\}^{M_i \times N_i}$, with $M_i < N_i$. This yields the following formulation via the Kronecker product:
\begin{equation}\label{eq_scs}
    \mathbf{y}_{\rm{s}}=(\mathbf{S}_1\otimes \mathbf{S}_2\otimes \cdots \otimes \mathbf{S}_q)\mathbf{y}=\mathbf{Sy}.
\end{equation}
This reduces the problem of choosing $M_\Pi$ out of $N_\Pi$ samples into $q$ problems where $M_i$ out of $N_i$ must be chosen. In general, $(M_\Sigma = \sum_{i=1}^q M_i) < M_\Pi$ and so the structured subsampling problem is simpler than the original one. However, the single hyperparameter $M_\Pi$ is replaced with $q$ hyperparameters $M_i$. In the next section, we elaborate on how a sampling budget is distributed across the axis, i.e. how we set all $M_i$ in SCOS.




%% file: rough_ara.tex
\section{Automatic Resource Allocation}\label{sec_ara}

\noindent Given a total sampling budget $M_\Sigma=\sum_{i=1}^{q}M_i$ and the number of dimensions $q$ in the data, the samples can be allocated in $\binom{M_{\sum} - 1}{q - 1}$ number of ways, resulting in a combinatorial resource allocation problem. Instead of heuristically specifying the active elements $M_i$ per axis, we automatically learn how to distribute the sampling budget. We achieve this by modifying the ML-based subsampling approach proposed in \cite{huijben2020DPS} so that it automatically performs resource allocation while learning which samples to preserve.

\begin{figure*}[ht!]
	\centering
	\includegraphics[width=0.8\textwidth, trim = {0cm 0cm 0cm 0.1cm}, clip]{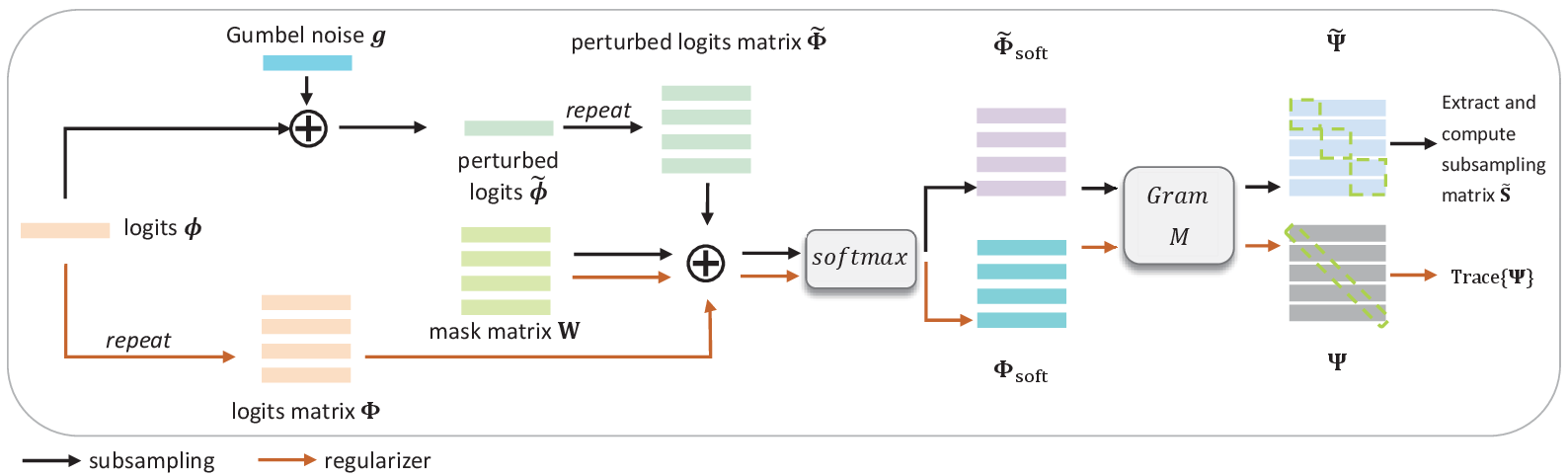}
	\caption{Flowchart of automatic resource allocation in SCOSARA. There are two parallel ways to obtain the subsampling matrices and the regularizer: the top path perturbs the logits with Gumbel noise to approximate the process of sampling from a categorical distribution, while the lower path directly uses the logits to produce the regularizer.}
	\label{fig_scosara}
    \vspace{-0.5cm}
\end{figure*}

In \cite{huijben2020DPS}, the design of subsampling matrices is treated as sampling without replacement from a categorical distribution. The log probabilities, or logits, of the categorical distribution are learned in a gradient-based fashion by employing the straight-through Gumbel estimator~\cite{jang2017categorical, maddison2017concrete}. In the forward direction, the Gumbel-max trick is employed, i.e. Gumbel noise $~\text{Gumbel}(0,1)$ is added to the logits, followed by the application of $\text{argmax}$ to sample from the distribution over elements in the subsampling matrix. During backpropagation, `soft samples' are drawn from the Gumbel-softmax distribution by relaxing the non-differentiable $\text{argmax}$ function with a $\text{softmax}$, allowing the usage of backpropagation for the optimization of the logits.

Based on this procedure, the subsampling matrices $\mathbf{S}_i$ can be learned based only on the total budget $M_\Sigma$, without specifying each $M_i$ separately. To this end, we introduce a single vector $\bm{\phi}$ containing $N_\Sigma=\sum_{i=1}^{q}N_i$ logits. The entries of $\bm{\phi}$ are ordered in the same order the dimensions of $\mathbf{Y}$ are vectorized in. The Gumbel-softmax trick is then used to obtain a differentiable approximation of the process of sampling $M_\Sigma$ items without replacement out of the total $N_\Sigma$. To do so, Gumbel noise $\bm{g} \in \real^{N_\Sigma}$, $\bm{g} \sim \text{Gumbel(0,1)}$ is added to the logits $\bm{\phi}$, yielding the perturbed logits vector $\tilde{\bm{\phi}} \in \real^{N_\Sigma}$. Next, the softmax function is applied to $\tilde{\bm{\phi}}$. To obtain $M_\Sigma$ samples without replacement, the largest entry of $\tilde{\bm{\phi}}$ is replaced by $-\infty$ (i.e. the probability of choosing the same entry again is set to 0), and then softmax is applied again~\cite{huijben2020DPS}. This process is repeated until an auxiliary matrix $\tilde{\bm{\Phi}}$ of size $M_\Sigma \times N_\Sigma$ is obtained. {The same effect can be achieved through the repetition of the perturbed logits vector $\tilde{\bm{\phi}}$ $M_\Sigma$ times, followed by the addition of a \emph{masking matrix} $\mathbf{W}$ of size $M_\Sigma \times N_\Sigma$ whose entries are taken from $\{-\infty, 0\}$.}

Since the entries of the logits vector $\bm{\phi}$ are ordered according to the vectorization of $\mathbf{Y}$, the auxiliary matrix $\tilde{\bm{\Phi}}$ exhibits the structure
\begin{align}
	\label{eq_phihat}
	\tilde{\bm{\Phi}}=
	\begin{bmatrix}
		\mathbf{S}_1 & \mathbf{N}_{1,2} & \cdots & \mathbf{N}_{1,q} \\
		\mathbf{N}_{2,1} & \mathbf{S}_2 & \cdots & \mathbf{N}_{2,q} \\
        \vdots & \vdots & \ddots & \vdots \\
		\mathbf{N}_{q,1} & \mathbf{N}_{q,2} & \cdots & \mathbf{S}_q
	\end{bmatrix}.
\end{align}
The subsampling matrices $\mathbf{S}_i \in \real^{M_i \times N_i}$ appear along the diagonal blocks of $\tilde{\bm{\Phi}}$, whereas the off-diagonal blocks $\mathbf{N}_{i,j} \in \real^{M_i \times N_j}, 1 \leq i,j \leq q$, contain only nuisance terms that are nonzero due to the usage of softmax.

Two crucial observations must be made regarding $\tilde{\bm{\Phi}}$. First, the sizes of the diagonal blocks are unknown because the $M_i$ are unknown. Second, since the entries of $\tilde{\bm{\phi}}$ change when the noise $\mathbf{g}$ is added and when $\bm{\phi}$ is modified through backpropagation, the matrix $\tilde{\bm{\Phi}}$ is not available in practice. Instead, the order the samples are drawn in, is unknown and only $\bm{\Pi} \tilde{\bm{\Phi}}$ can be obtained, where $\bm{\Pi} \in \mathbb{R}^{M_\Sigma \times M_\Sigma}$ is an unknown permutation matrix. Both problems are addressed by computing $\tilde{\bm{\Psi}} = (\bm{\Pi} \tilde{\bm{\Phi}})^\mathrm{T} (\bm{\Pi} \tilde{\bm{\Phi}}) = \tilde{\bm{\Phi}}\trans \tilde{\bm{\Phi}} \in \mathbb{R}^{N_\Sigma \times N_\Sigma}$, yielding a matrix of the general form
\begin{align}
	\label{eq_psi}
	\tilde{\bm{\Psi}}=
	\begin{bmatrix}
		\mathbf{S}_1\trans \mathbf{S}_1 + \mathbf{N}'_{1}  & \mathbf{N}'_{1,2} & \cdots & \mathbf{N}'_{1,q} \\
		\mathbf{N}'_{2,1} & \mathbf{S}_2\trans \mathbf{S}_2 + \mathbf{N}'_{2} & \cdots & \mathbf{N}'_{2,q} \\
        \vdots & \vdots & \ddots & \vdots \\
		\mathbf{N}'_{q,1} & \mathbf{N}'_{q,2} & \cdots & \mathbf{S}_q\trans \mathbf{S}_q + \mathbf{N}'_{q}
	\end{bmatrix}.
\end{align}
The block matrices along the diagonal of $\tilde{\bm{\Psi}}$, i.e. $\tilde{\mathbf{S}}_i = \mathbf{S}_i^\mathrm{T} \mathbf{S}_i + \mathbf{N}'_{i} \in \real^{N_i \times N_i}$, can then be used to construct
\begin{align}
\vspace{-0.4cm}
	\label{eq_subsamp}
	\tilde{\mathbf{S}} = \tilde{\mathbf{S}}_1 \otimes \tilde{\mathbf{S}}_2 \otimes \cdots \otimes \tilde{\mathbf{S}}_q \in \real^{N_\Pi \times N_\Pi}.
\vspace{-0.4cm}
\end{align}

The structure of $\tilde{\bm{\Psi}}$ also motivates the following observation. Consider the unperturbed logits $\bm{\phi}$ and apply the Gumbel-softmax trick without the noise $\mathbf{g}$ so as to obtain a new auxiliary matrix $\bm{\Phi}$, and from it, $\bm{\Psi} = \bm{\Phi}\trans \bm{\Phi}$. Since the rows of ${\bm{\Phi}}$ are non-negative and sum to one due to the $\softmax$ function, maximizing $\mathrm{Trace}(\bm{\Psi})$ (equivalently, $\|{\bm{\Phi}}\|_\mathrm{F}^2$) reduces the magnitude of the off-diagonal elements of $\bm{\Psi}$, thereby aligning the forward and backward passes of the straight-through estimator. This term can thus be used as a regularizer when optimizing the logits.

The overall procedure described throughout this section is illustrated in Figure \ref{fig_scosara}. The upper branch of the figure uses Gumbel noise $\mathbf{g}$ to construct the matrix $\tilde{\bm{\Psi}}$ used for sampling, while the lower branch is noiseless and yields the regularizing matrix $\bm{\Psi}$.

%% file: sec_optimization_target.tex
\section{Optimization Target}\label{sec_optimization_target}

\noindent Replacing task-based optimization because of its time and resource-intensive nature, we take FIM and CRB into account.
The CRB is a lower bound for the variance of unbiased estimators \cite{kay1993fundamentals}. However, CS methods are biased, the sources of the bias being amplitude errors \cite{austin2013dynamic}, parameter discretization errors \cite{chi2011sensitivity}, and the enforcing of sparse support \cite{ben2010cramer}. In spite of this, the cited works illustrate that the CRB correctly describes the behavior of practical CS estimators when the signal-to-noise ratio is high and the sampling grid is fine enough.

The CRB is computed as the inverse of FIM, which quantifies the amount of information that an observable random variable carries about the parameters of its distribution. The FIM is also closely related to the Restricted Isometry Property (RIP) \cite{blanchard2011compressed}. The inter-column coherence of a model matrix is inversely proportional to the eigenvalues of the FIM \cite{ben2010cramer,austin2009relation}. Similarly, the Restricted Isometry Constant (RIC) involved in the RIP is smaller when the eigenvalues of the FIM are large.

Therefore, instead of the downstream task such as directly computing the signal recovery, we try to maximize the trace of FIM for two reasons. First, it largely reduces the network and training complexity since it is estimator-agnostic. Second, the A-optimality criterion, which refers to the minimization of the trace of the CRB, is a commonly employed optimization target in the design of experiments. However, in the present setting, directly using the CRB would require multiple matrix inversions during training. The maximization of the trace of FIM can be seen as a heuristic for the CRB, as it can be formulated in terms of the harmonic mean of the eigenvalues of the CRB matrix and is therefore closely related to A-optimality. 

In the derivation of the cost function, we let $\mathbf{Ax}$ obey a differentiable parametric model $\mathbf{b}$ with a real-valued parameter vector $\bm{\xi}$, so that the subsampled data $\mathbf{y}_{\rm{s}}$ can be reformulated as:
\begin{equation}\label{eq_fim_1}
\vspace{-0.2cm}
    \mathbf{y}_{\rm{s}} = \mathbf{S(Ax+n)} = \mathbf{S(b(\bm{\xi})+n)} = \bm{\mu}+\tilde{\mathbf{n}},
\end{equation}
where $\mathbf{b}$ denotes the noiseless fully sampled data and $\bm{\mu}$ corresponds to the noiseless compressed data. The noise $\mathbf{n}$ is circularly-symmetric AWGN characterized by the distribution $\mathcal{CN}(\boldsymbol{0}, \boldsymbol{\Sigma})$ and $\boldsymbol{\Sigma}=\sigma^2\mathbf{I}$. Instead, since we can only compute $\tilde{\mathbf{S}}$ from SCOSARA, following the notation in (\ref{eq_subsamp}), we rewrite this as:
\begin{equation}\label{eq_fim_2}
\vspace{-0.2cm}
    \mathbf{y}_{\rm{s}} = \tilde{\mathbf{S}}\mathbf{(b(\bm{\xi})+n)} =  \tilde{\bm{\mu}}+\tilde{\mathbf{n}},
\end{equation}
where the covariance of $\tilde{\mathbf{n}}$ is approximately given by $\sigma^2 \mathbf{SS}\trans$. Based on the Slepian-Bangs formalism, the FIM can be written as 
\begin{align} \label{eq_fim}
    \bm{\mathcal{J}}& \approx \frac{2}{\sigma^2}\mathfrak{R} \left\{ \left( \frac{\partial \bm{\mu}}{\partial \bm{\xi}}\right)\herm \left( \frac{\partial \bm{\mu}}{\partial \bm{\xi}}\right) \right\} 
    = \frac{2}{\sigma^2}\mathfrak{R} \left\{ \bm{J}_b\herm \tilde{\mathbf{S}}\trans \tilde{\mathbf{S}} \bm{J}_b \right\}
\end{align}
where $(\cdot)\herm$ is the conjugate transpose, $\mathfrak{R}$ refers to extracting the real-valued part and $\bm{J}_b$ stands for the Jacobian matrix of data $\mathbf{b}$ with respect to parameters $\bm{\xi}$. Note that in the case when multiple targets coexist, the CRB of a single target can also adequately describe the estimation problem when they are well-spaced. In summary, the FIM-driven SCOSARA aims to maximize Fisher information, resulting in the following cost function:
\begin{equation}\label{eq_cost_function}
    \mathcal{L}=\argmin_{\bm{\phi}}\,-\mathrm{Trace}(\bm{\mathcal{J}})-\mathrm{Trace}( \bm{\Psi}).
\end{equation}

%% file: sec_evaluation.tex
\section{Evaluation}\label{sec_evaluation}

\noindent As an illustrative example of multidimensional data acquisition, we detail the application of the SCOS  within the context of multichannel ultrasound localization. The simulation scenarios and experimental settings are similar to \cite{wangicassp2024}, but we use a larger data model whose parameters are summarized in TABLE \ref{tab_hyperparameters}. The task is to localize a single scatterer in the Region of Interest (ROI) whose parameter vector can be formulated by $\bm{\xi}=[x, z, a, \varphi]$, which corresponds to its two-dimensional coordinates, reflectivity coefficient, and phase.
The dataset is generated by assuming a single scatterer with a varying reflectivity coefficient is randomly located in a given ROI. 
\begin{table}[t]
\vspace{-0.2cm}
\caption{Model parameters summary\label{tab_hyperparameters}}
\centering
\begin{tabular}{ccccccc}
\toprule
\textbf{Parameter} & $N_{\rm{T}}$ & $N_{\rm{R}}$ & $N_{\rm{F}}$ & $N_{\sum}$ & $N_{\Pi}$ & $M_{\sum}$\\
\midrule
\textbf{Value} & $64$ & $64$ & $113$ & $241$ & $462848$ & $30 \sim 220$\\
\bottomrule
\end{tabular}
\vspace{-0.5cm}
\end{table}

We design/learn three subsampling matrices for reducing transmitters, receivers, and Fourier coefficients, denoted $\mathbf{S}_{\rm{T}}\in \real^{M_{\rm{T}}\times N_{\rm{T}}}$, $\mathbf{S}_{\rm{R}}\in \real^{M_{\rm{R}}\times N_{\rm{R}}}$, and $\mathbf{S}_{\rm{F}}\in \real^{M_{\rm{F}}\times N_{\rm{F}}}$. 
They compress all three axes while retaining as much information about scatterer locations as possible. 
To demonstrate the performance of \textbf{SCOSARA}, we compare it against \textbf{Uniform Compression} and other SOTA algorithms, including ML-based methods \textbf{DPS-topK}~\cite{huijben2020learning} and \textbf{J-DPS}~\cite{wangicassp2024}, and a \textbf{Greedy Search} algorithm. To provide comprehensive and fair comparisons, we consider the following points in the evaluation:
\begin{itemize}
    \item We trained SCOSARA first and then used its resource allocation scheme for other algorithms for two reasons: the baseline methods provide no resource allocation scheme, and the compression factor ${M_{\Pi}}/{N_{\Pi}}$ should be consistent. 
    \item We test $20$ values of $M_{\sum}$ from $30$ to $220$ at intervals of $10$, each has a distinct compression factor.
\end{itemize}

\noindent \textbf{Baselines:} The Uniform Compression scheme implies the equally spaced allocation of $M_{\rm{T}}$ choices among $N_{\rm{T}}$, and so forth. 
The DPS-TopK refers to optimizing a vector of logits $\bm{\phi}_\Pi\in \real^{N_\Pi}$ and using the Gumbel-$\topK$ trick in the forward pass to select $M_\Sigma$ elements set to $1$. 
The J-DPS divides the long vector into three smaller dimension-specific ones.
The Greedy Search algorithm iteratively discards the entry of $\bm{\phi}$ which reduces $\rm{Trace}(\mathcal{J})$ the least until the desired compression factor is achieved.

\begin{figure}[t]
	\centering
    \vspace{-0.3cm}
	\includegraphics[width=1\columnwidth, trim = {0cm 0.1cm 0cm 0.5cm}, clip]{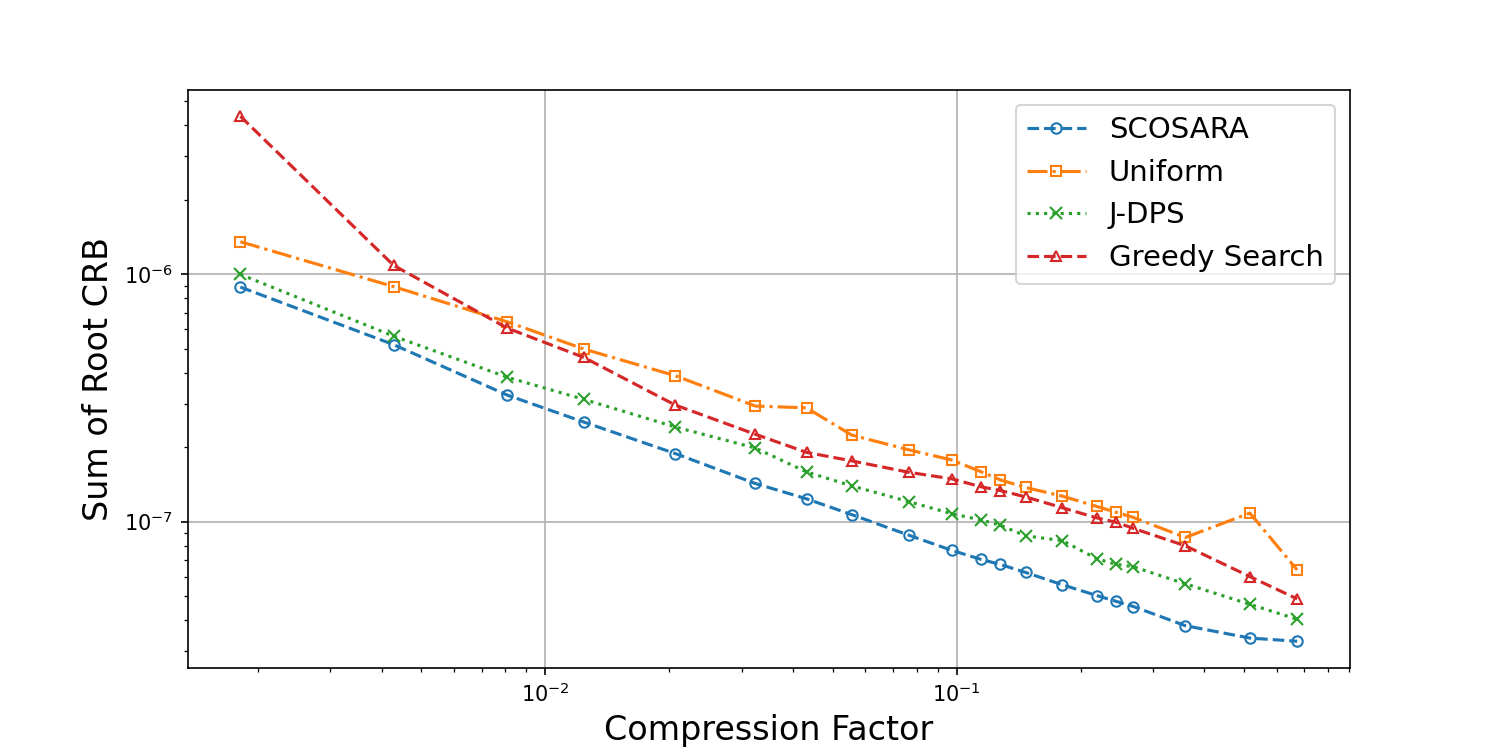}
	\caption{CRB as a function of the compression factor (higher factor denotes more selected samples). }
    \vspace{-0.5cm}
	\label{fig_crb_vs_ratio}
\end{figure}

\noindent \textbf{Results:}  We used an NVIDIA A100 GPU node to run all experiments, the DPS-TopK failed to operate due to computational constraints because it optimizes a large logits vector $\bm{\phi}_\Pi$, which requires the generation of large matrices of shape $(M_{\Pi}\times N_{\Pi})$ in the computational graph. 
After running other algorithms given all the values of $M_{\sum}$, at each compression factor, we apply the four SCOS schemes to the evaluation dataset and compute the CRB values based on their compressed data. The resulting four curves are illustrated in Fig. \ref{fig_crb_vs_ratio}, showing that SCOSARA outperforms other methods, achieving the lowest CRB across all compression factors.

\begin{figure}[t]
    \centering
    \begin{subfigure}[b]{0.45\linewidth}
        \centering
        \includegraphics[width=\linewidth]{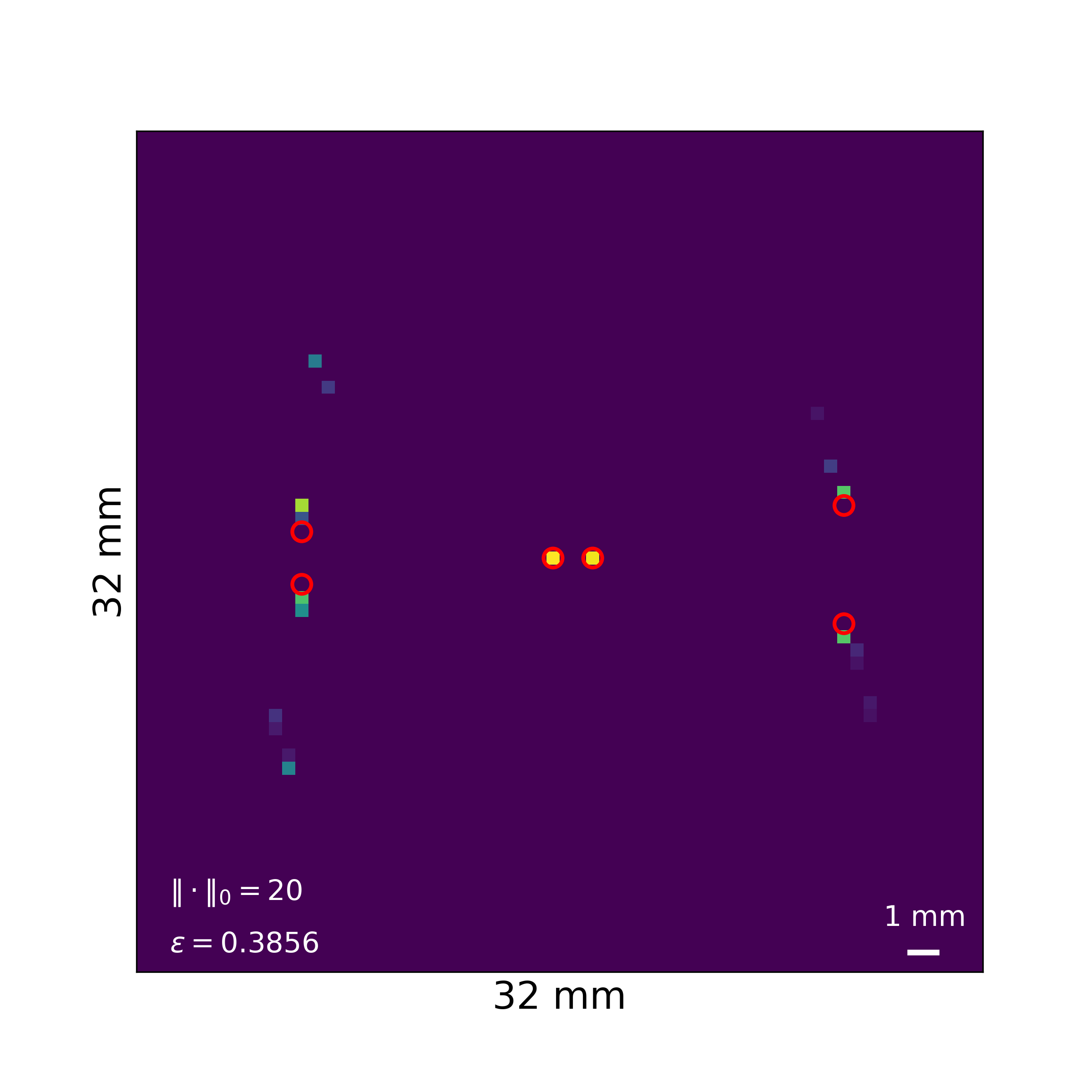}
        \caption{Uniform compression}
        \label{fig:sub5}
    \end{subfigure}
    \hfill
    \begin{subfigure}[b]{0.45\linewidth}
        \centering
        \includegraphics[width=\linewidth]{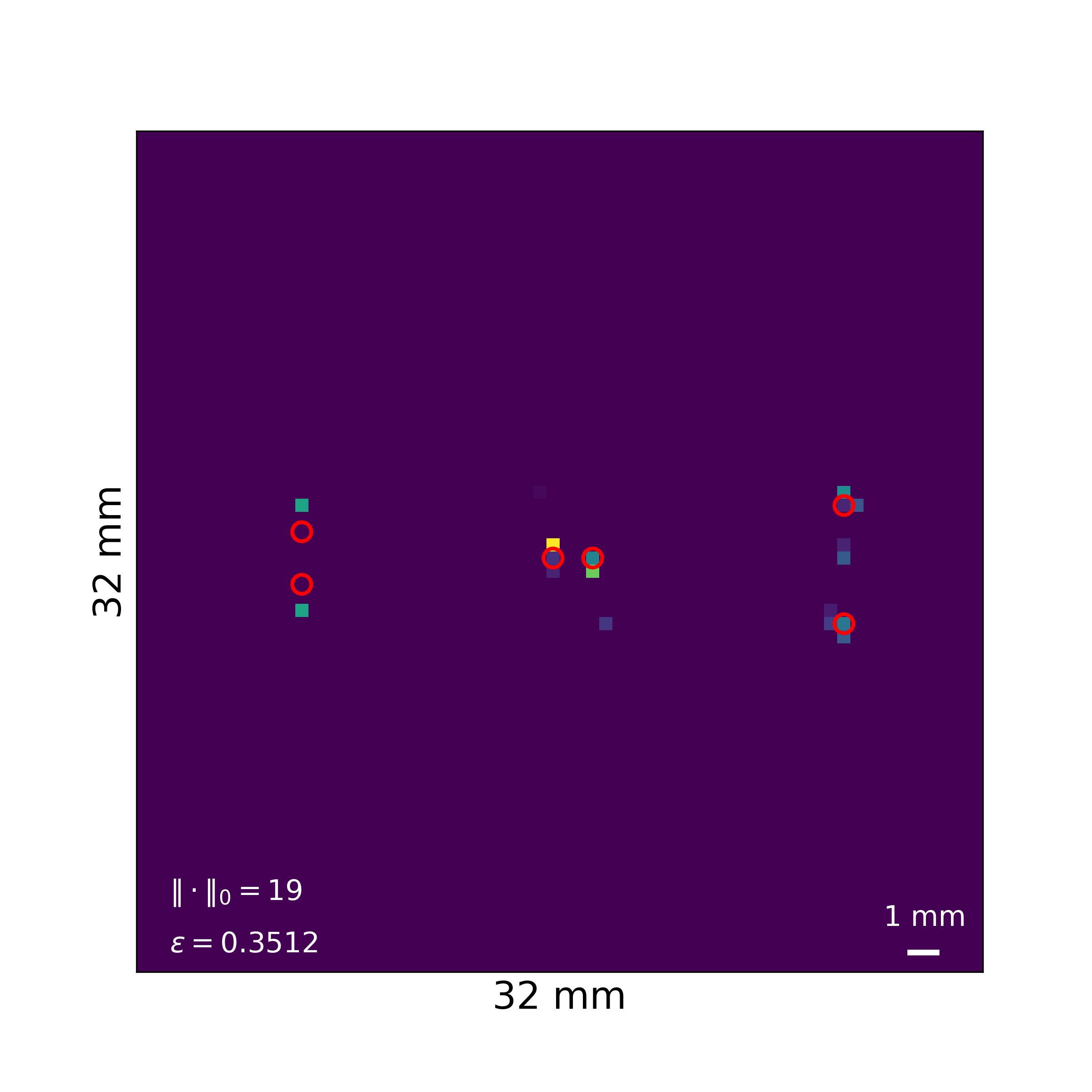}
        \caption{Greedy search}
        \label{fig:sub6}
    \end{subfigure}
    \begin{subfigure}[b]{0.45\linewidth}
        \centering
        \includegraphics[width=\linewidth]{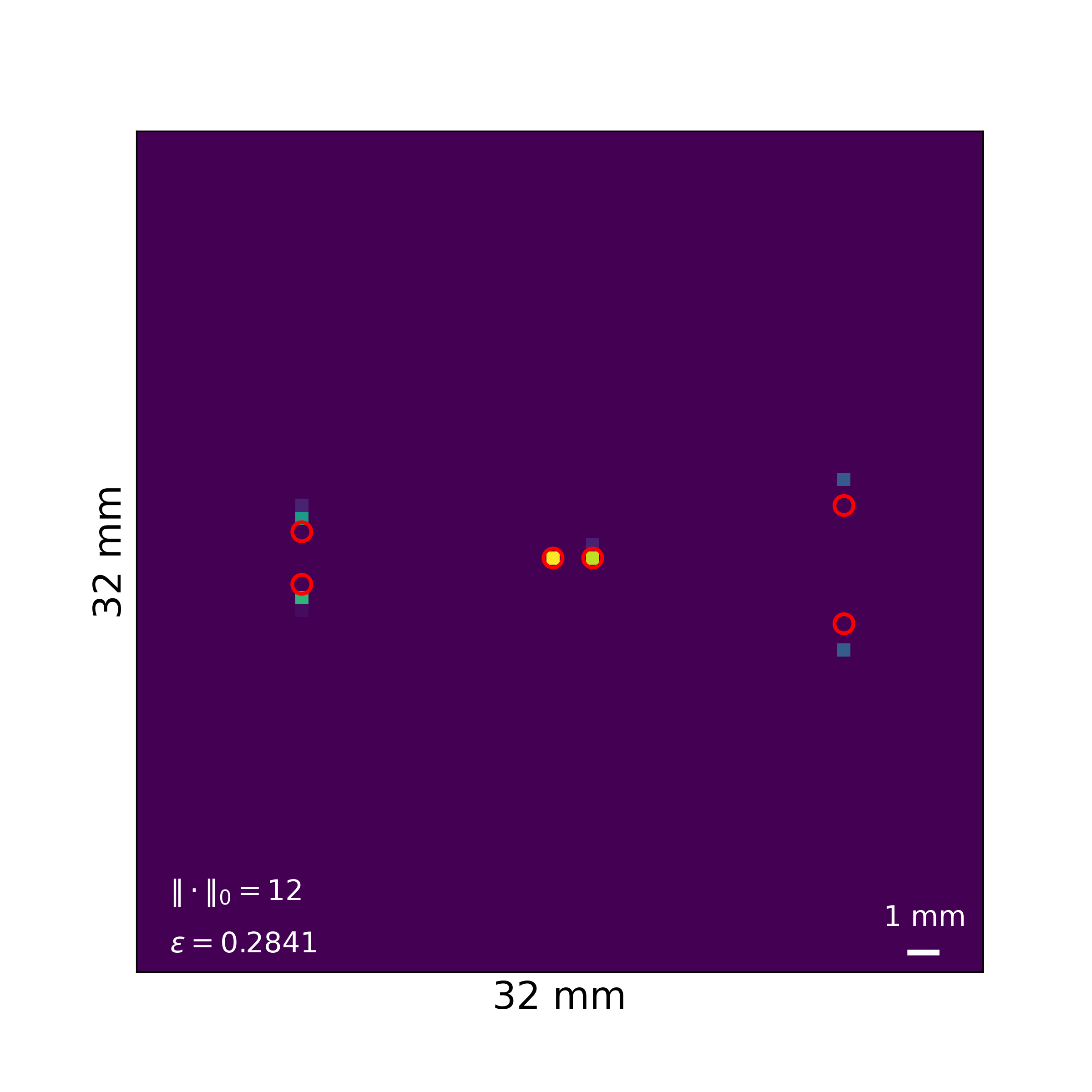}
        \caption{J-DPS}
        \label{fig:sub7}
    \end{subfigure}
    \hfill
    \begin{subfigure}[b]{0.45\linewidth}
        \centering
        \includegraphics[width=\linewidth]{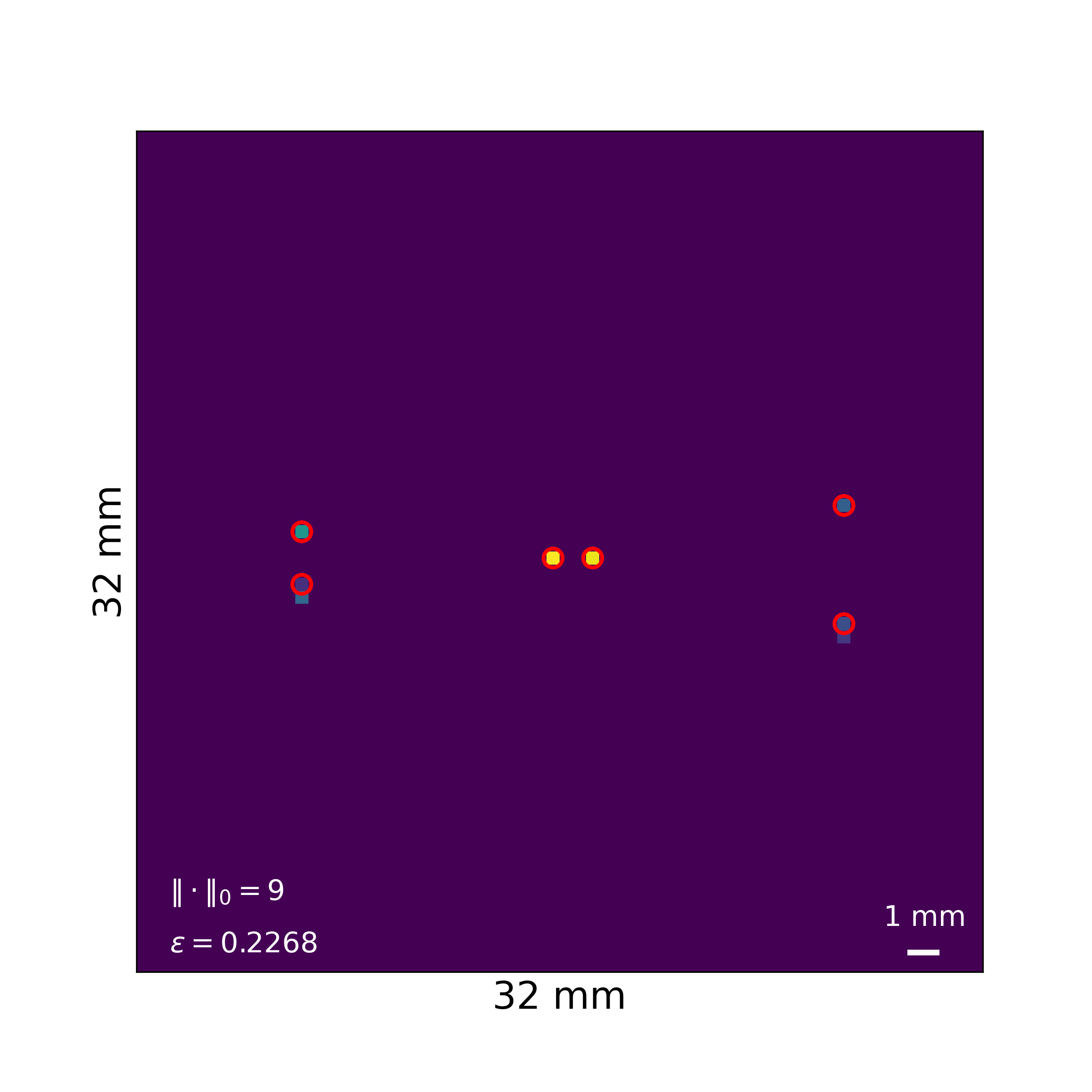}
        \caption{SCOSARA}
        \label{fig:sub8}
    \end{subfigure}
    \caption{Reconstructed images comparison in the multi-scatterer cases when $M_{\sum}=120$. The ground truth scatterers are represented by red circles. }
    \label{fig_reco_compare}
    \vspace{-0.5cm}
\end{figure}

Furthermore, we also evaluate the recovery performance in the multi-scatterer case. Knowing the distance of two pixels is approximately equal to half the wavelength in the simulation, we define three pairs of scatterers whose distances are two, three, and four pixels, respectively. We applied the four SCOS schemes given $M_{\sum}=120$ along with the $2000$-iteration complex FISTA, obtaining the reconstructed images. We computed the MSE $\epsilon$ and the number of nonzero elements $\Vert \cdot \Vert_0$ for quantitative comparison. As illustrated in Fig. \ref{fig_reco_compare}, SCOSARA outperforms with respect to image quality and localization accuracy.
 
%% file: sec_conclusion.tex
\vspace{-0.1cm}
\section{Conclusion}\label{sec_conclusion}

In this work, we introduce the concept of FIM-driven SCOSARA, which can automatically allocate the sampling resources to each dimension while maximizing the Fisher information. Using ultrasound multichannel localization as a case study, we quantitatively compare SCOSARA with other baseline algorithms, where it shows superior performance in terms of CRB analysis and recovery performance. Preliminary results not included in this work show that the SCOSARA can be extended to allow preferential compression of user-specified axes by replacing the regularization term $\text{Trace}\{\bm{\Psi}\}$ with $\text{Trace}\{\mathbf{D} \bm{\Psi}\}$, where $\mathbf{D}$ is a diagonal matrix. These results will be presented along with experiments on measurement data and comparisons against task-based algorithms by involving convolutional sparse coding algorithms~\cite{wang2024eusipco} in a future manuscript.
